\documentclass[journal,twoside,web]{ieeecolor}
\usepackage{generic}
\usepackage{cite}
\usepackage{amsmath,amssymb,amsfonts}
\usepackage{algorithmic}
\usepackage{graphicx}
\usepackage{algorithm,algorithmic}
\usepackage{hyperref}
\hypersetup{hidelinks=true}
\usepackage{textcomp}
\usepackage{booktabs}
\usepackage{eso-pic}

\def\BibTeX{{\rm B\kern-.05em{\sc i\kern-.025em b}\kern-.08em
    T\kern-.1667em\lower.7ex\hbox{E}\kern-.125emX}}
\begin{document}
\title{Adaptive-CaRe: Adaptive Causal Regularization for Robust Outcome Prediction}
\author{Nithya Bhasker, Fiona R. Kolbinger, Susu Hu, Gitta Kutyniok, Stefanie Speidel \IEEEmembership{Member, IEEE}
\thanks{Nithya Bhasker, Susu Hu and Stefanie speidel are with the Department of Translational Surgical Oncology, National Center for Tumor Diseases (NCT), NCT/UCC Dresden, a partnership between DKFZ, Faculty of Medicine and University Hospital Carl Gustav Carus, TUD Dresden University of Technology, and Helmholtz-Zentrum Dresden-Rossendorf (HZDR), Germany.
(e-mail: nithya.bhasker@nct-dresden.de, stefanie.speidel@nct-dresden.de).}
\thanks{Fiona R. Kolbinger is with Weldon School of Biomedical Engineering, Purdue University, West Lafayette, IN, USA; and  Department of Visceral, Thoracic and Vascular Surgery, University Hospital and
Faculty of Medicine Carl Gustav Carus, TUD Dresden University of Technology, Germany.}
\thanks{Gitta Kutyniok is with Ludwig-Maximilians-Universität München,
Munich, Germany; Munich Center for Machine Learning (MCML), Munich, Germany; University of Tromsø, Tromsø,
Norway; DLR-German Aerospace Center, Germany.}}

\maketitle

\AddToShipoutPictureBG*{%
  \AtPageLowerLeft{%
    \parbox[b]{\paperwidth}{\centering\footnotesize
       \vspace{0.5cm} 
       This work has been submitted to the IEEE for possible publication. Copyright may be transferred without notice, after which this version may no longer be accessible.
       \vspace{0.8cm} 
    }%
  }%
}

\begin{abstract}
Accurate prediction of outcomes is crucial for clinical decision-making and personalized patient care. Supervised machine learning algorithms, which are commonly used for outcome prediction in the medical domain, optimize for predictive accuracy, which can result in models latching onto spurious correlations instead of robust predictors. Causal structure learning methods on the other hand have the potential to provide robust predictors for the target, but can be too conservative because of algorithmic and data assumptions, resulting in loss of diagnostic precision. Therefore, we propose a novel model-agnostic regularization strategy, Adaptive-CaRe, for generalized outcome prediction in the medical domain. Adaptive-CaRe strikes a balance between both predictive value and causal robustness by incorporating a penalty that is proportional to the difference between the estimated statistical contribution and estimated causal contribution of the input features for model predictions. Our experiments on synthetic data establish the efficacy of the proposed Adaptive-CaRe regularizer in finding robust predictors for the target while maintaining competitive predictive accuracy. With experiments on a standard causal benchmark, we provide a blueprint for navigating the trade-off between predictive accuracy and causal robustness by tweaking the regularization strength, $\lambda$. Validation using real-world dataset confirms that the results translate to practical, real-domain settings. Therefore, Adaptive-CaRe provides a simple yet effective solution to the long-standing trade-off between predictive accuracy and causal robustness in the medical domain. Future work would involve studying alternate causal structure learning frameworks and complex classification models to provide deeper insights at a larger scale. \textit{The code will be made available upon publication.} 
\end{abstract}


\section{Introduction}
\label{sec:introduction}
Outcome prediction in the healthcare domain involves the modelling of routinely collected patient data using statistical or machine learning methods to predict clinical events of interest such as disease progression, treatment response and patient survival. Accurate prediction of outcomes can enhance clinical decision-making and personalized patient care, ultimately improving patient outcomes \cite{rajkomar2019machine,adams2022prospective}. Given its potential for significant impact, outcome prediction has emerged as a critical area of focus in both clinical research and practice.

However, predictive models in the healthcare domain frequently face challenges with generalization, that is, maintaining predictive accuracy when applied to different patient populations, clinical settings, or institutions than those they were trained on \cite{futoma2020myth}. Limited generalization primarily stems from variations in patient demographics, differences in clinical practices, inconsistent data collection standards, and changes over time \cite{ghassemi2020review,yang2024generalizability}. Consequently, many models demonstrate high accuracy in their training contexts but fail when deployed externally, significantly restricting their practical utility in clinic.

A well-known example is the proprietary Epic Sepsis Model, deployed widely in emergency departments across the United States of America.  External validation studies have reported poor performance of the model on retrospective patient cohorts. In one case, the model missed 67\% of the cases with sepsis, despite generating frequent false alarms \cite{wong2021external}. In another, the model demonstrated poor sensitivity, performing worse than a random alert system \cite{ostermayer2024external}. These failures illustrate how models that rely on superficial correlations and subtle site-specific artefacts rarely survive real-world domain shifts.

Existing work either employ data-based or capacity-based regularization techniques to improve generalization. Capacity-based regularization techniques \cite{tibshirani1996regression, hoerl1970ridge, hinton2012improving, srivastava2014dropout, goodfellow2016deep, krogh1991simple} add an explicit penalty that shrinks or constrains the hypothesis space, while data-based regularization techniques \cite{dao2019kernel, zhang2018mixup, yun2019cutmix, frid2018gan} enrich the training distribution so the model fits many plausible variants, rather than memorising the originals. However, most of these models are still susceptible to exploiting surface correlations instead of underlying causal mechanisms. 

Causal structure learning serves as a powerful alternative to addressing these generalization challenges in outcome prediction \cite{prosperi2020causal}. By identifying causal relationships rather than mere statistical correlations, causal structure learning facilitates the development of robust and interpretable predictive models that generalize beyond their initial training contexts \cite{pearl2009causal, scholkopf2022causality}. However, prioritizing causal structure alone can lead to the exclusion of non-causal, yet highly informative, biomarkers essential for maximizing diagnostic precision \cite{rothenhausler2021anchor}.

To overcome these limitations, we introduce \textit{\textbf{Adaptive-CaRe}}, an \textbf{Adaptive} \textbf{Ca}usal \textbf{Re}gularization strategy, for generalized outcome prediction in the medical domain. Adaptive-CaRe strikes a balance between predictive value and causal robustness by incorporating an adaptive penalty that is proportional to the difference between the estimated statistical contribution and the estimated causal contribution of the input features for model predictions. Our contributions are: 
\begin{itemize}
    \item We propose a novel model-agnostic regularization strategy, Adaptive-CaRe, for generalized outcome prediction in the medical domain.
    \item We delve into the working mechanism of the regularization framework by considering a simple MLP model and synthetic data with known causal graph. 
    \item By considering a standard causal benchmark, we provide proof-of-concept for the flexibility of the model to strike a balance between predictive value and causal robustness. 
    \item We also validate the method on standard real-world datasets and provide insights about the predictions of the model after employing Adaptive-CaRe penalty.
\end{itemize}

\section{Related work}

\subsection*{Regularization techniques}
Capacity-based regularization techniques, such as L1 and L2 penalties \cite{tibshirani1996regression, hoerl1970ridge}, dropout \cite{hinton2012improving, srivastava2014dropout}, weight decay \cite{krogh1991simple}, and early stopping \cite{goodfellow2016deep} introduce explicit constraints on model parameters to reduce the effective capacity of the hypothesis space. These methods aim to improve generalization by discouraging overly complex solutions: the L2 penalty or the weight decay encourage small, diffuse parameter values \cite{hoerl1970ridge}, while the L1
penalty promotes sparsity by driving some parameters exactly to zero \cite{tibshirani1996regression}. Dropout takes a different approach by randomly omitting units during training, implicitly constraining the model’s capacity and reducing reliance on specific features \cite{hinton2012improving, srivastava2014dropout}. Collectively, these strategies limit model complexity and are grounded in statistical learning theory, which links reduced hypothesis space capacity with improved generalization performance, especially in over-parametrized models \cite{vapnik2013nature}.

Data-based regularization techniques improve generalization by enriching the training distribution with plausible variations of the input data, encouraging models to learn invariant features rather than memorising specific instances \cite{dao2019kernel, zhang2018mixup}. These methods augment data through transformations, synthesis, or domain-specific perturbations, thereby increasing coverage of the input space. For example, \cite{yun2019cutmix} introduced CutMix, a strategy that replaces random patches of an image with patches from another training image, forcing the model to identify objects based on partial views, and acting as a strong regularizer that boosts robustness against input corruptions. Similarly, \cite{frid2018gan} demonstrated the efficacy of leveraging Generative Adversarial Networks (GANs) to synthetically augment liver lesion datasets, significantly improving classification performance on limited medical data. Such strategies act as implicit regularizers by aligning training conditions more closely with real-world variability.

Despite their widespread use, both capacity-based and data-based regularizers have critical limitations. Capacity-based methods like dropout and weight decay \cite{srivastava2014dropout, hinton2012improving, krogh1991simple} act as black-box constraints, suppressing model complexity without leveraging domain knowledge, which can inadvertently limit useful expressiveness. Data-based techniques, such as augmentation or synthetic sampling \cite{yun2019cutmix, frid2018gan}, often rely on superficial correlations (e.g., rotations, noise) that do not reflect causal structures, leading to failures under distribution shifts \cite{geirhos2020shortcut}. Moreover, both approaches offer limited interpretability and control, with studies showing that augmented training can still overfit spurious features or dataset artefacts \cite{recht2019imagenet}.

\subsection*{Causal Structure Learning}
Most causal structure learning methods aim to infer causal relationships from observational data within the framework of causal graphical models \cite{pearl2009causal}. These approaches are typically grouped into three main categories: (i) constraint-based methods, such as the PC or FCI algorithms, which rely on conditional independence tests to recover the causal graph or equivalence class under assumptions such as faithfulness, with some (e.g., PC) requiring causal sufficiency and others (e.g., FCI) accounting for latent confounders \cite{spirtes2000causation, kalisch2007estimating}; (ii) functional causal model-based methods, which leverage assumptions about the functional form and noise structure (e.g., additive noise or linear non-Gaussianity) to identify causal directions \cite{shimizu2006linear, hoyer2009nonlinear}; and (iii) score-based methods, which search over possible graph structures using scoring criteria and optimization strategies, including greedy algorithms \cite{chickering2002optimal} and continuous relaxations for differentiable structure learning \cite{zheng2018dags, ng2022masked}.

Causal structure learning methods, while valuable for uncovering underlying mechanisms, can be too rigid or assumption-heavy to perform well in practical prediction tasks. Their focus on identifiability and robustness under interventions often comes at the cost of predictive accuracy \cite{scholkopf2012causal}, especially when assumptions like causal sufficiency or faithfulness are violated \cite{montagna2023assumption, kalisch2007estimating}. Moreover, causal discovery algorithms can be unstable in high-dimensional or noisy settings, limiting their utility in data-driven applications \cite{heinze2018causal}. These limitations motivate hybrid approaches that integrate the robustness and interpretability of causal models with the predictive power of statistical learning, aiming to leverage the strengths of both paradigms.

\subsection*{Hybrid causal-predictive networks}
Ge \textit{et al}. \cite{geinvariant} propose an Invariant Structure Learning (ISL) framework that jointly learns a causal graph and predictive model by enforcing that the predictive relationships remain invariant across different data environments. This approach integrates structure learning with robust optimization to improve generalization under distributional shifts and to recover interpretable causal structures. Their method demonstrates improved performance on synthetic and real-world tabular datasets. However, applying ISL to healthcare domain remains challenging due to difficulties in defining meaningful environments, small subgroup sizes, computational complexity, and the dynamic nature of clinical data.

Kyono \textit{et al}. \cite{kyono2020castle} introduce a causal structure learning based regularizer, CASTLE, for improving generalization in supervised learning. They add a supervised loss term to the non-linear framework from \cite{zheng2018dags} to learn the target variable. CASTLE \cite{kyono2020castle} is one of the revolutionary works which demonstrated the superior performance of causality based regularization over commonly used regularization techniques for deep learning such as L1-norm, L2-norm, dropout and early stopping \cite{tibshirani1996regression, hoerl1970ridge, goodfellow2016deep}. In healthcare settings, however, its application is impeded by high model complexity relative to dataset size, scalability challenges in high-dimensional spaces, and limited clinical interpretability of learned graphs. Additionally, CASTLE couples the prediction task with an auxiliary reconstruction objective, forcing the model to reconstruct features based on their causal parents. This coupling can hinder outcome prediction accuracy when the learned structure is imperfect or too rigid for complex clinical data.

We address these research gaps by disentangling the causal structure learning from the prediction pipeline. We also adopt adaptive penalty that is proportional to the difference between the statistical contribution and causal contribution of the input features. The adaptive penalty provides the flexibility to optimize for both predictive accuracy and causal robustness. 

\section{Empirical Risk Minimization}
In a standard supervised learning setting, we consider a domain comprising of the input space $\mathcal{X} \subseteq \mathbb{R}^d$ and the target space $\mathcal{Y}$. We are provided with a dataset $\mathcal{D} = \{(\mathbf{x}_i, y_i)\}_{i=1}^N$ consisting of $N$ independent and identically distributed (i.i.d.) samples drawn from an unknown joint probability distribution $P_{XY}$ over random variables $X$ and $Y$.

The objective is to learn a hypothesis $f_\theta: \mathcal{X} \to \mathcal{Y}$, parameterized by $\theta \in \Theta$, that minimizes the expected risk $R(\theta) = \mathbb{E}_{(X, Y) \sim P_{XY}}[\mathcal{L}(Y, f_\theta(X))]$. Since $P_{XY}$ is unknown, we approximate the expected risk via Empirical Risk Minimization (ERM) \cite{shalev2014understanding}. Specifically, we seek the optimal parameters $\hat{\theta}$ that minimize the regularized empirical loss:

\begin{equation}
    \hat{\theta} = \underset{\theta \in \Theta}{\mathrm{argmin}} \left[ \frac{1}{N} \sum_{i=1}^N \mathcal{L}(y_i, f_\theta(\mathbf{x}_i)) + \lambda \Omega(\theta) \right]
    \label{eq:objective_function}
\end{equation}

where $\mathcal{L}: \mathcal{Y} \times \mathcal{Y} \to \mathbb{R}_{\geq 0}$ denotes the loss function (e.g., cross-entropy), $\Omega(\theta)$ is a regularization term, and $\lambda \geq 0$ is a hyperparameter controlling regularization strength.

\section{Adaptive-CaRe}
In the medical domain, we apply the supervised learning setup to problems such as disease classification and outcome prediction, where the goal is to predict a clinical outcome $Y$ given biological markers $X$. However, medical data rarely satisfies the strict i.i.d.\ assumptions. Variations in patient demographics or acquisition protocols introduce distribution shifts, leading to scenarios where the training distribution differs from the test distribution, denoted as $P_{\text{train}}(X, Y) \neq P_{\text{test}}(X, Y)$. Under these conditions, standard ERM models (e.g., Multilayer Perceptrons) are prone to exploiting spurious correlations, i.e., statistical associations present in $P_{\text{train}}$ that do not hold in $P_{\text{test}}$.

\begin{figure}
    \centering
    \includegraphics[width=1\linewidth]{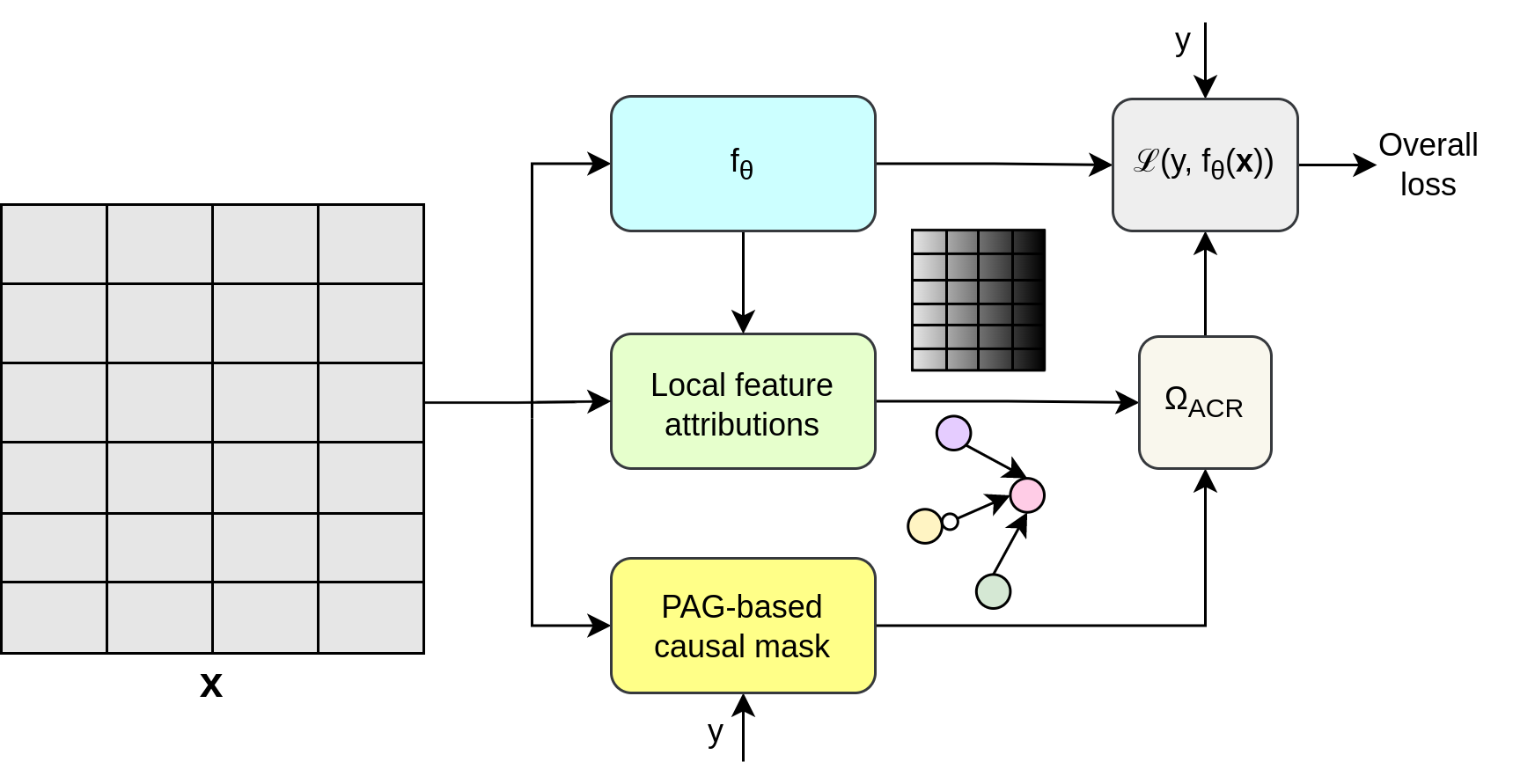}
    \caption{\textbf{Adaptive-CaRe} strategy: Features whose statistical contribution to the prediction is significantly different from their estimated causal contribution are penalized.}
    \label{fig:block_diagram}
\end{figure}

To mitigate this, we propose \textit{Adaptive-CaRe}, an Adaptive Causal Regularization strategy, for generalized outcome prediction in the medical domain. Adaptive-CaRe strikes a balance between predictive value and causal robustness by incorporating an adaptive penalty that is proportional to the difference between the estimated statistical contribution and the estimated causal contribution of the input features. As illustrated in Fig.~\ref{fig:block_diagram}, this involves: (i) Estimation of Partial Ancestral Graph (PAG)-based causal mask, (ii) Computation of Local Feature Attributions, and (iii) Calculation of Adaptive-CaRe penalty. 

\subsection{PAG-based causal mask}
Many existing causal structure learning algorithms rely on stringent assumptions, such as causal sufficiency and acyclicity \cite{kalisch2007estimating, zheng2018dags}. However, observational biomedical data frequently violate these premises due to the prevalence of unmeasured confounding and selection bias. Consequently, we employ the \textit{Fast Causal Inference (FCI)} algorithm \cite{spirtes2000causation} to identify a subset of robust predictors. Unlike algorithms that assume causal sufficiency (e.g., PC), FCI is asymptotically correct even in the presence of latent variables and selection bias.

We utilize FCI to estimate a Partial Ancestral Graph (PAG), denoted as $\mathcal{G}$, over the observed variables. From $\mathcal{G}$, we compute a binary adjacency mask $A \in \{0, 1\}^d$ derived from the subset of features $S \subset \{1, \dots, d\}$ corresponding to the \textit{Robust Predictors} of the target variable $Y$. In the context of a PAG, an edge implies a specific structural relationship based on its endpoints. We define the set $S$ by selecting variables connected to $Y$ via:
\begin{enumerate}
    \item \textbf{Directed Edges ($X \rightarrow Y$):} Representing an ancestral relationship where $X$ is a cause of $Y$ (tail at $X$, arrowhead at $Y$).
    \item \textbf{Partially Directed Edges ($X \circ\!\rightarrow Y$):} Representing a possible ancestral relationship where $X$ is not an effect of $Y$, though unmeasured confounding may exist (circle at $X$, arrowhead at $Y$).
    \item \textbf{Bidirected Edges ($X \leftrightarrow Y$):} Representing the presence of an unmeasured confounder affecting both $X$ and $Y$, where neither variable causes the other (arrowhead at both $X$ and $Y$). While not a direct cause, $X$ serves as a proxy for the unobserved confounder, improving predictive accuracy.
\end{enumerate}

We employ FCI to identify the causal structure of the data. Our reliance on the resulting causal parents for robust prediction is grounded in the \textit{Assumption of Invariant Mechanisms}, which posits that the causal mechanism generating the target variable is stable across environments. Formally, while the marginal distribution of features $P(X)$ and the conditional distributions of non-causal features may vary across environments $e \in \mathcal{E}$, we assume the conditional distribution of the target given its selected predictors, $P(Y \mid X_S)$, remains effectively invariant:

\begin{equation}
    P^e(Y \mid X_S) = P^{e'}(Y \mid X_S) \quad \forall e, e' \in \mathcal{E}
    \label{eq:invariance}
\end{equation}

In reality, the environments $\mathcal{E}$ might correspond to distinct data acquisition protocols, variations in measuring devices, or different clinical centers. The implementation details are elaborated in Appendix~\ref{fci_details}. 

\subsection{Local Feature Attributions}
Next, we compute the \textit{Gradient $\times$ Input} attributions \cite{shrikumar2017learning}, which estimates the contribution of each input feature to the final prediction by combining local sensitivity with the feature's magnitude. The attribution score $S_j$ for the $j$-th feature $x_j$ is computed as the partial derivative of the output with respect to $j$-th input feature, weighted by the input feature's value:

\begin{equation}
    S_j = x_j\frac{\partial f(\mathbf{x})}{\partial x_j}
\end{equation}

To identify the most influential predictors regardless of the direction of their effect, we consider the absolute value of the attribution score as the final feature importance value, $|S_j|$. 

\subsection{Adaptive-CaRe penalty}
Given the PAG-based causal mask and the local feature attributions, we penalize the model's reliance on non-causal features by applying a penalty that is the sum of the feature attributions for the non-causal features:

\begin{equation}
    \Omega_{ACR} = \frac{1}{N} \sum_{i=1}^{N} \sum_{j=1}^{d} \left| S_{ij} \right| \, (1 - A_j)
    \label{eq:AdaCare}
\end{equation}

We can rewrite Equation \ref{eq:objective_function} as:

\begin{equation}
    \hat{\theta} = \underset{\theta \in \Theta}{\mathrm{argmin}} \left[ \frac{1}{N} \sum_{i=1}^N \mathcal{L}(y_i, f_\theta(\mathbf{x}_i)) + \lambda \Omega_{ACR} \right]
    \label{eq:objective_function_ACR}
\end{equation}

Fundamentally, standard supervised learning minimizes the \textit{observational risk} over all features, $\mathbb{E}[\mathcal{L}(Y, f(X))]$, which guarantees generalization only under i.i.d.\ conditions. In contrast, by penalizing the model's statistical reliance on non-causal features, we allow the predictor $f_\theta$ to target stable causal mechanisms rather than transient statistical dependencies, thereby effectively relaxing the strict i.i.d.\ requirement and improving generalization on unseen data. The predictor $f_\theta$ could be any state-of-the-art classification method.  

There are often cases in the medical domain, where it is not known \textit{a priori} if there exist causal mechanisms between the biological markers measured and the target variable. These cases are generally aimed at instant triaging by utilizing the easy predictors. We account for these cases, by tweaking the regularization strength parameter $\lambda$. We discuss the influence of this parameter in the later sections.  

\section{Synthetic Data Generation}
\label{sec:synthetic_data}
To assess the model's robustness to spurious correlations and distribution shifts, we generated a synthetic dataset governed by a known data-generating process. The data consists of $N$ samples, each comprising five continuous features.

The observed feature set includes two causal parents ($X_1, X_2$), one proxy feature ($X_{\text{proxy}}$), one spurious feature ($X_{\text{spurious}}$), and one random noise feature ($X_{\text{noise}}$). These are designed to simulate complex interactions, selection bias, and observational noise:

\begin{itemize}
    \item Causal Parents ($X_1, X_2$): Continuous features sampled from a standard normal distribution.
    \item Target ($Y$): A binary label generated via a non-linear logistic model dependent on an interaction between $X_1$ and $X_2$.
    \item Proxy Feature ($X_{\text{proxy}}$): A feature that is a noisy descendant of the target $Y$ (representing reverse causality or a downstream symptom).
    \item Spurious Feature ($X_{\text{spurious}}$): A feature that exhibits a strong correlation with $Y$ in the training environment due to artificial selection bias, but becomes uncorrelated noise in the testing environment.
    \item Noise ($X_{\text{noise}}$): An uninformative Gaussian noise feature.
\end{itemize}

The structural equations governing the data generation are defined as follows. First, the causal parents and uninformative noise are sampled:
\begin{equation}
    X_1, X_2, X_{\text{noise}} \sim \mathcal{N}(0, 1)
\end{equation}

The binary target $Y$ is sampled from a Bernoulli distribution, $Y \sim \text{Bernoulli}(\pi)$, where the probability $\pi$ is determined by a sigmoid function $\sigma(\cdot)$ applied to a non-linear interaction term:
\begin{equation}
    \pi = \sigma\left(1.5 X_1 + 1.5 X_2 + 2.0 (X_1 X_2)\right)
\end{equation}

Crucially, the proxy and spurious features are generated conditional on the label $Y$, creating a correlation structure distinct from causal ancestry. Let $\mu_y = 2.0$ if $Y=1$ and $\mu_y = -2.0$ if $Y=0$. The downstream features are sampled as:
\begin{align}
    X_{\text{proxy}} &\sim \mathcal{N}(\mu_y, 1) \\
    X_{\text{spurious}} &\sim \begin{cases} 
        \mathcal{N}(\mu_y, 1) & \text{if mode} = \text{train} \\
        \mathcal{N}(0, 9) & \text{if mode} = \text{test}
    \end{cases}
\end{align}

In the training environment, $X_{\text{spurious}}$ is a strong predictor of $Y$ (mimicking a "shortcut" signal), whereas in the testing environment, it represents high-variance noise uncorrelated with the target.

\paragraph*{\textbf{Medical Analogy}} This setup simulates predicting a clinical outcome ($Y$) based on gold-standard continuous biomarkers ($X_1, X_2$) that interact non-linearly. $X_{\text{proxy}}$ represents a symptom caused by the outcome (e.g., fever). $X_{\text{spurious}}$ represents a confounding procedural artifact that is strongly correlated with disease status in the training data due to selection bias but is absent or random in real-world deployment.

\begin{figure}
    \centering
    \includegraphics[width=1.0\linewidth]{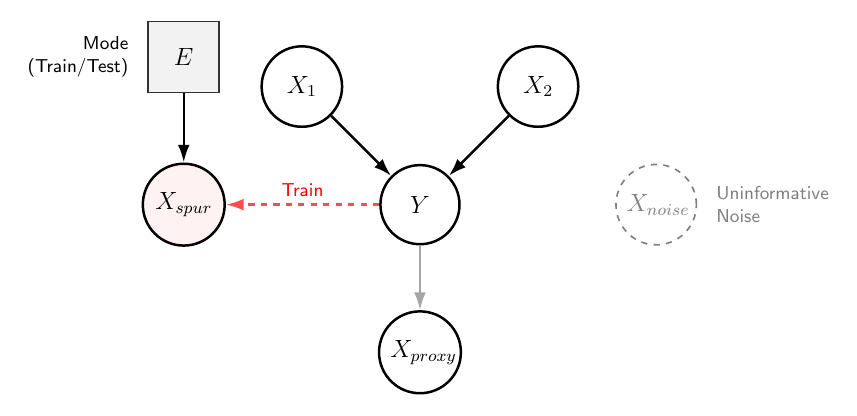}
    \caption{The ground truth causal graph for the generated synthetic data. Variables: $X_{1,2}$ - causal parents, $X_{proxy}$ - proxy, $X_{spur}$- spurious feature, $E$ - environment variable, $X_{noise}$- noise, $Y$ - target}
    \label{fig:synthetic_data}
\end{figure}

\section{Results}
The results are organized in the following manner: (i) We delve into the working mechanism of Adaptive-CaRe (ACR) with help of synthetic data with known causal mechanisms (Section~\ref{sec:synthetic_data}, (ii) By considering a standard causal benchmark, we demonstrate the flexibility of the regularization framework in optimizing for both predictive value and causal robustness, (iii) Finally, we validate the framework on real-world dataset in the medical domain and provide insights about the prediction results after employing ACR penalty. 

\paragraph{Experimental setup} We generate $1000$ train and $1000$ test samples using three different seeds for the experiments on synthetic data, unless otherwise specified. For the experiments with causal benchmark and the real-world dataset, we employ five-fold cross-validation to split the datasets into five folds of train and test samples. All experiments tackle the classification problem. While the ACR regularizer, $\Omega_{ACR}$, is model-agnostic and compatible with any state-of-the-art predictor, $f_\theta$, we prioritized simplicity to isolate the contribution of our regularization strategy without the confounding overhead of complex architectures. Therefore, we employed two basic models incorporated with ACR penalty: a two-layer Multilayer Perceptron (MLP) and Logistic Regression (LR). 

For standard baselines, we choose the scikit-learn implementation of Logistic Regression (LR), and a two-layer MLP without regularization. We also compare against the regularized baselines by applying weight decay (WD) and early stopping (ES) to the MLP model. As state-of-the-art causal regularizer, we choose CASTLE \cite{kyono2020castle}. We also compare the results with a LR model after selecting only the features provided by the PAG-based causal mask (LR w/ causal FS).

All the models are trained for a maximum of $1000$ iterations. All the MLP models are two-layer MLP implemented in Pytorch and consist of one hidden layer with 32 neurons, followed by ReLU activation, and an output layer. LR w/ ACR is implemented in Pytorch for the ease of applying custom penalty and consists of one linear layer. The MLP models are initialized with random weights sampled from Xavier normal distribution. We choose Adam optimizer with a weight decay of $1e^-5$ (if opted). For the models with early stopping, we stop training after $100$ iterations if the training loss does not change by $>1e^-5$ for 30 iterations. We report precision, recall, and F1-score evaluation metrics for all the experiments. We also compute and plot  SHAP values-based feature importances for LR, MLP w/ WD and ES, and MLP w/ ACR models. The details about this are provided in Appendix~\ref{shap}.  We also display the features identified by the PAG-based mask alongside the feature importance scores for reference. As the weighted DAG estimated by CASTLE is not comparable to the SHAP-based values computed for the other models, we restrict our analysis to the above mentioned models.  

\subsection{Generalization in the presence of proxy and spurious features}
We generate data from Section~\ref{sec:synthetic_data} using three different seeds to assess the performance of MLP and LR models with Adaptive-CaRe regularizer (ACR). We report the median F1 score metrics for train and test data generated using the three different seeds. The results are provided in Table~\ref{tab:synth_exp1}.

\begin{table}[t]
    \centering
    \caption{Generalization in the presence of spurious features.}
    \label{tab:synth_exp1}
   \begin{tabular}{lcc}
    \toprule
    \textbf{Model} & \textbf{F1-score (train)} & \textbf{F1-score (test)} \\
    \midrule
    LR                  & 0.99 [0.99, 0.99] & 0.76 [0.75, 0.76] \\
    MLP                 & 0.99 [0.99, 0.99] & 0.81 [0.80, 0.84] \\
    MLP w/ WD           & 0.99 [0.99, 0.99] & 0.76 [0.76, 0.78] \\
    MLP w/ WD and ES    & 0.99 [0.99, 0.99] & 0.78 [0.77, 0.81] \\
    LR w/ causal FS     & 0.75 [0.75, 0.76] & 0.75 [0.74, 0.76] \\
    CASTLE              & 0.64 [0.61, 0.64] & 0.56 [0.55, 0.57] \\
    \midrule
    \textbf{LR w/ ACR}  & 0.68 [0.49, 0.83] & 0.71 [0.47, 0.83] \\
    \textbf{MLP w/ ACR} & 0.89 [0.88, 0.92] & 0.81 [0.80, 0.84] \\
    \bottomrule
    \end{tabular}
\end{table}

\begin{figure}
    \centering
    \includegraphics[width=1\linewidth]{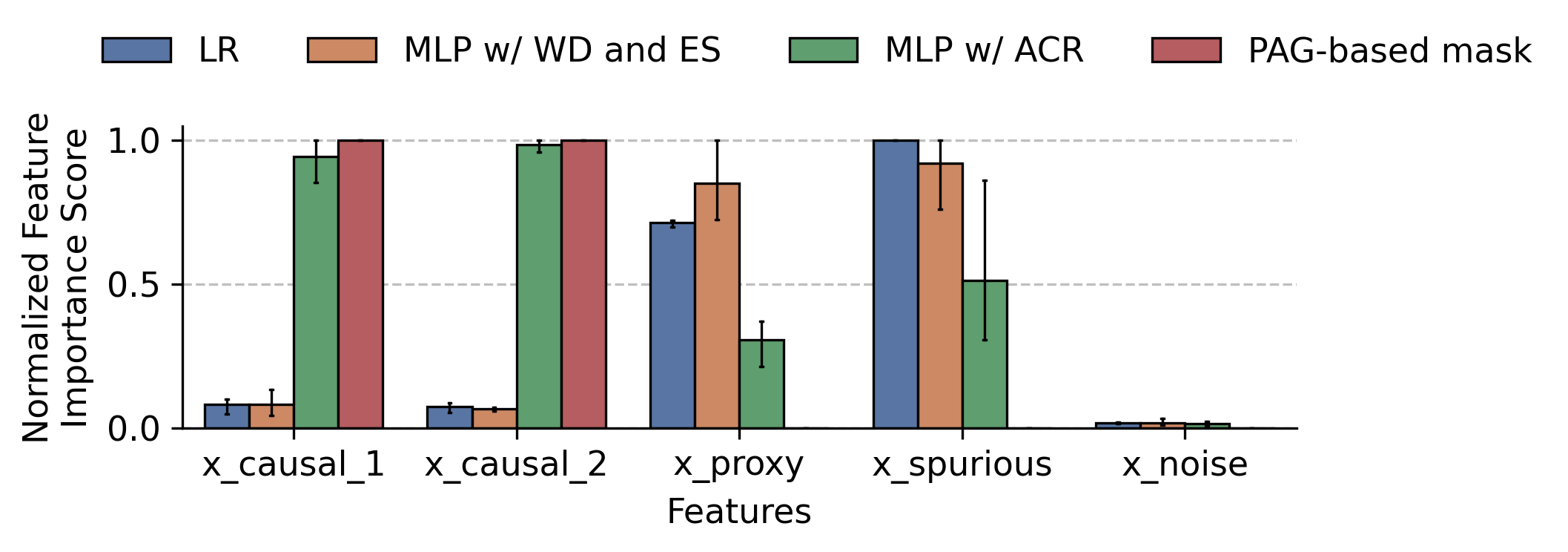}
    \caption{Illustration of normalized feature importance scores for the synthetic dataset.}
    \label{fig:feat_imp_syn}
\end{figure}

Models regularized with Adaptive-CaRe outperform CASTLE, the causal regularization baseline, on test data. This could be due to the strict assumptions about causal sufficiency made by the NOTEARS framework \cite{zheng2018dags}, upon which CASTLE is developed. The LR with causal FS model, although trained on only the causal parents of the target, does not outperform the  MLP w/ ACR model, indicating rigidity to optimize for predictive accuracy. The significant difference between the train and test results for the LR and MLP variants without Adaptive-CaRe regularizer, suggests that these models might be relying on spurious features. To verify this hypothesis, we plotted the normalized feature importance scores for LR, MLP w/ WD and ES, and MLP w/ ACR. We also display the features identified by the PAG-based mask. As seen in Fig.~\ref{fig:feat_imp_syn}, MLP w/ ACR is successful in suppressing spurious, proxy, and noise features and place a higher importance on the causal parents. LR and MLP w/ WD and ES models, on the other hand, optimize only for predictive accuracy and place the spurious and proxy features ahead of the causal parents. These results reinforce the benefits of balancing causal robustness with predictive accuracy.

\subsection{Influence of Regularization strength}
\begin{figure}
    \centering
    \includegraphics[width=0.45\linewidth]{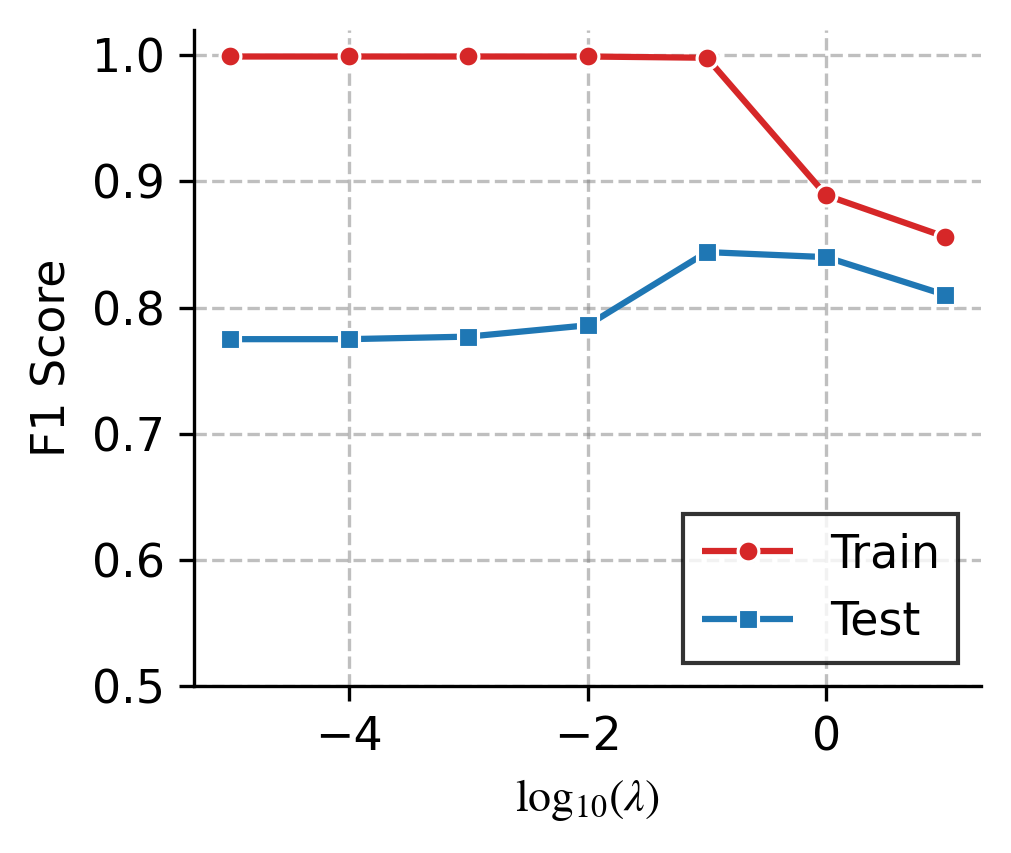}
    \includegraphics[width=0.45\linewidth]{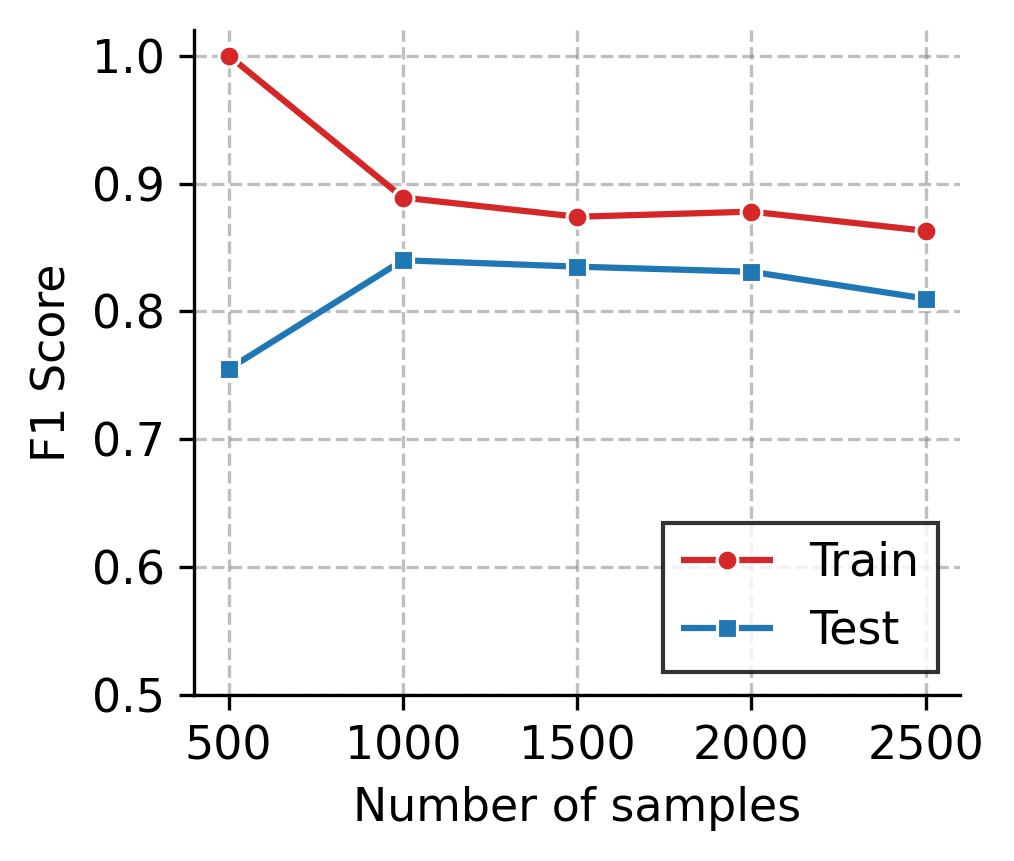}
    \caption{Influence of regularization strength $\lambda$ (left), and sample size $N$ (right) on the performance of MLP w/ Adaptive-CaRe. The line in red indicates train metrics, while the line in blue indicates test metrics.}
    \label{fig:synth_lambda}
\end{figure}

We study the influence of regularization strength parameter $\lambda$ for the MLP model with Adaptive-CaRe regularizer. The regularization strength controls the rigidity of the causal penalty. We make use of the synthetic dataset described in Section~\ref{sec:synthetic_data}. Fig~\ref{fig:synth_lambda} illustrates the train F1-score (red) and test F1-score (blue) of the model for different values of lambda: \{1e-5, 1e-4, 1e-3, 1e-2, 1e-1, 1, 10\}. 

\begin{figure}
    \centering
    \includegraphics[width=1\linewidth]{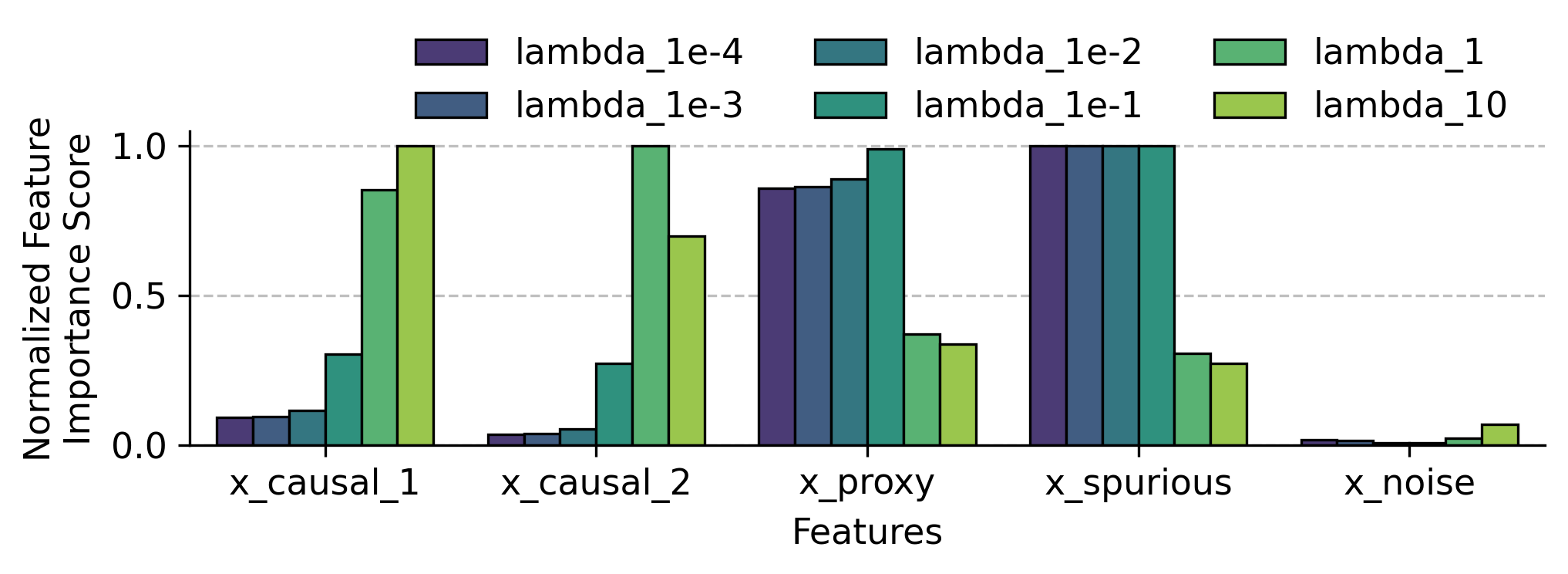}
    \caption{Influence of regularization strength $\lambda$. Increasing the regularization strength reduces the model's reliance on spurious features.}
    \label{fig:lambda_imp}
\end{figure}

The figure clearly demonstrates that increasing the regularization strength $\lambda$, encourages the model to close the gap between train and test F1-scores by relying on robust relationships in the features. To illustrate this, we plotted the normalized feature importance scores for the model. As demonstrated in Fig.~\ref{fig:lambda_imp}, increasing the regularization strength reduces the model's reliance on spurious features. Therefore, the results of this experiment provide a blueprint for the selection of the hyperparameter, $\lambda$.

\subsection{Influence of sample size}
We also study the influence of sample size on the performance of MLP model w/ Adaptive-CaRe regularizer by varying the sample size of synthetic data $N=\{500, 1000, 1500, 2000, 2500\}$. The results are provided in Fig.~\ref{fig:synth_lambda}. As observed in the figure, the model performance increases with increasing sample size until $N=1000$, and subsequently plateaus. This shows that the model trained with Adaptive-CaRe regularizer fares well in the low-data regime. 

\subsection{Predictive Value vs. Causal Robustness}
In this section, we demonstrate the trade off between predictive value and causal robustness by presenting a case study on diagnostic reasoning for hypertension (high BP). We utilize the standard Alarm network (A Logical Alarm Reduction Mechanism) \cite{lauritzen1988local}. This model captures the probabilistic dependencies between risk factors to provide an alarm message system for patient monitoring. 

\begin{figure*}
    \centering
    \includegraphics[width=1\linewidth]{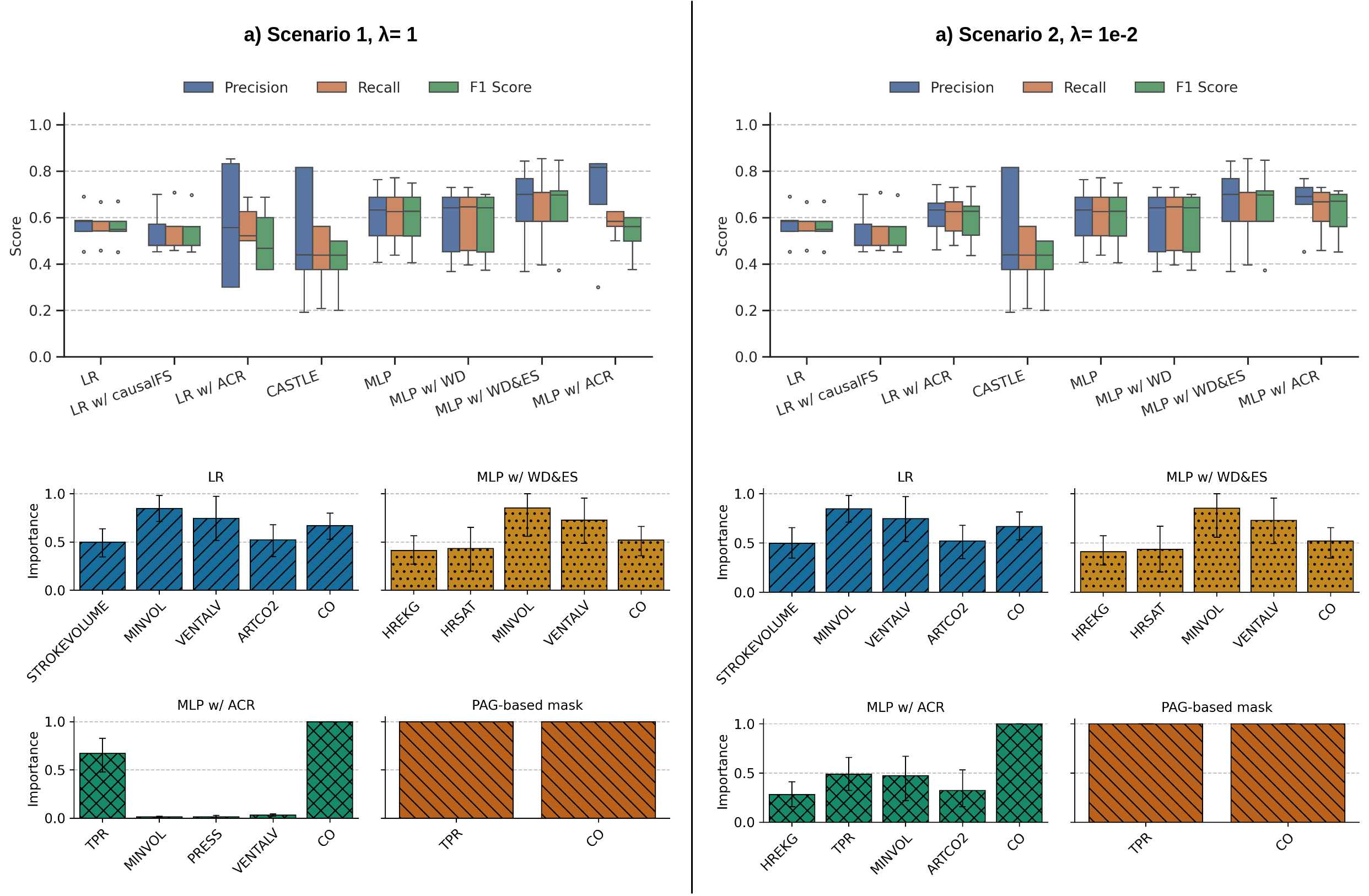}
    \caption{The performance of the models on the test splits of the Alarm dataset in different scenarios. The normalized feature importance scores for the top-5 features in both the scenarios are also illustrated.}
    \label{fig:alarm}
\end{figure*}

To simulate the low-data regime, we sample 100 instances from the ALARM network to predict Hypertension based on the \textit{BP} variable. The BP variable in the dataset contains 3 levels (High, Normal and Low). We binarize the variable for explanation simplicity. We consider the hypertension case (High BP: 1, and Low/Normal BP: 0). We employ a five fold cross validation scheme for the experiments. We present here two scenarios: 

\paragraph{Scenario 1} In this scenario, we optimize for causal robustness and set the regularization strength to $\lambda = 1.0$. The test results are presented in Fig.~\ref{fig:alarm}. Additionally, the normalized importance scores for the top-5 features for each model are also illustrated in Fig.~\ref{fig:alarm}. MLP w/ ACR (with a median test F1 score of $0.56$) performs better than a LR model (with a median test F1 score of $0.55$). MLP w/ WD and ES shows the best performance with a median test F1 score of $0.70$. However, MLP w/ ACR prioritizes the true causal parents CO (Cardiac output) and TPR (Total peripheral resistance), while the other models latch on to distant relatives or easy predictors. 

\paragraph{Scenario 2} In this scenario, we reduce the regularization strength to $\lambda = 1e^-2$ to optimize for both predictive accuracy and causal robustness. The test results and the normalized importance scores for the top-5 features for each model are illustrated in Fig.~\ref{fig:alarm}. In this scenario, MLP w/ ACR shows a significant increase in predictive accuracy with a median test F1 score of $0.67$ by picking easy predictors like MINVVOL (minimum volume), ARTCO2 (arterial $CO_2$), and HREKG (heart rate measured by an EKG monitor) in addition to the causal parents, CO and TPR.  

Therefore, Adaptive-CaRe strategy (ACR) provides the flexibility to either prioritize causal robustness, or strike a balance between both causal robustness and predictive accuracy depending on the requirements in the deployment environment by just tuning the regularization strength $\lambda$.

\subsection{Validation on SUPPORT dataset}
Next, we extend our validation to a real-world benchmark: the SUPPORT (Study to Understand Prognoses Preferences Outcomes and Risks of Treatment) dataset \cite{knaus1995support}. To simulate a low-data regime, we randomly subsampled 500 patients from the dataset\footnote{{\url{https://hbiostat.org/data/repo/supportdesc}}}. We address the binary classification problem of in-hospital mortality.

To ensure a realistic deployment setting and prevent data leakage, we rigorously excluded surrogate outcomes and administrative features containing future information. These exclusions comprise death time, total length of stay, physician scores (prg2m, prg6m), model predictions (surv2m, surv6m), composite scores (sps, aps), hospital charges (total and micro), average TISS score, and 'DNR' (Do Not Resuscitate) status and order dates. Missing values for physiological features were imputed using the ranges provided in the dataset documentation. Features with high missingness and no provided reference for imputation were removed to maintain data quality.

\begin{figure}
    \centering
    \includegraphics[width=1\linewidth]{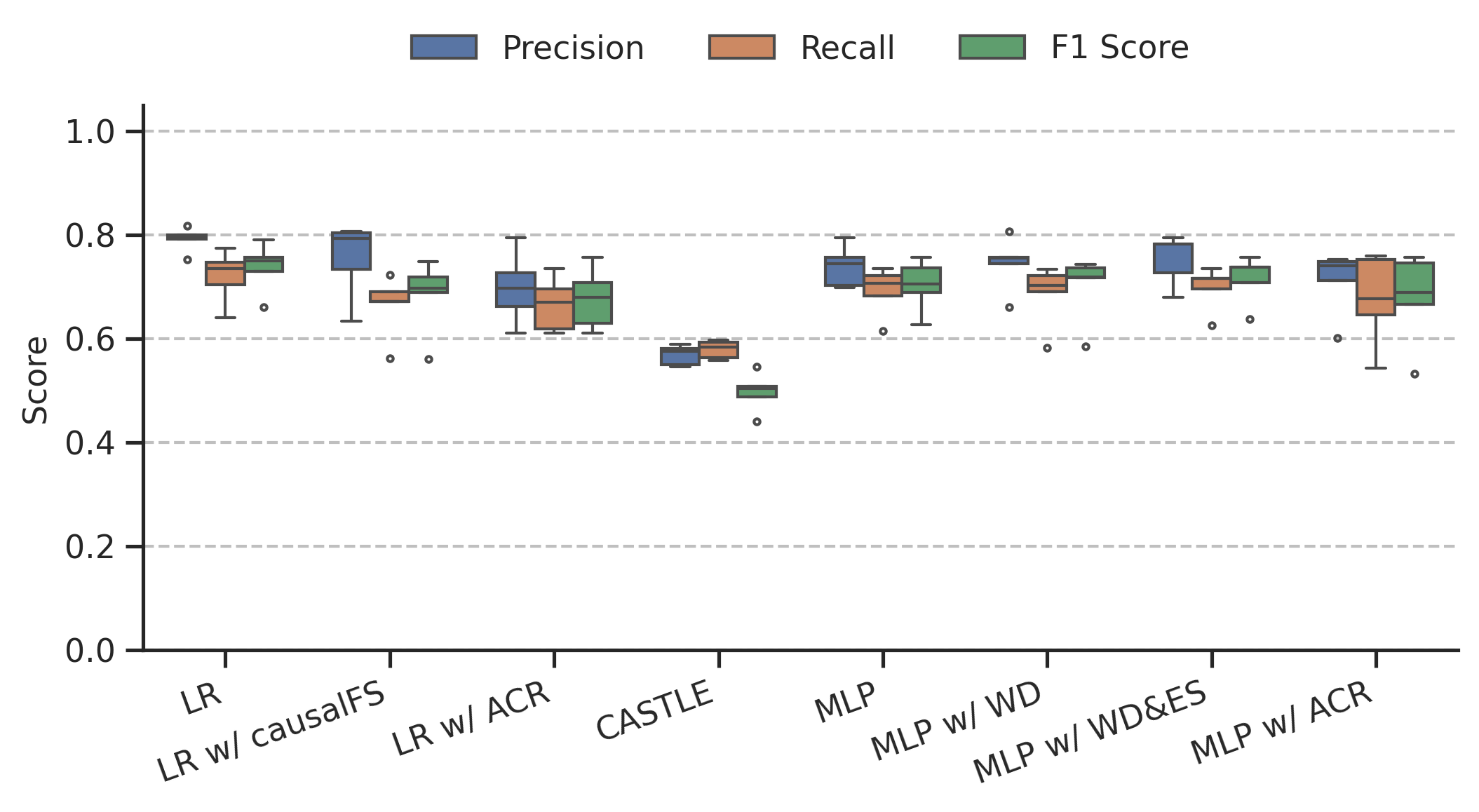}
    \caption{The performance of the models on the test splits of the SUPPORT dataset.}
    \label{fig:support_box}
\end{figure}

\begin{figure}
    \centering
    \includegraphics[width=1\linewidth]{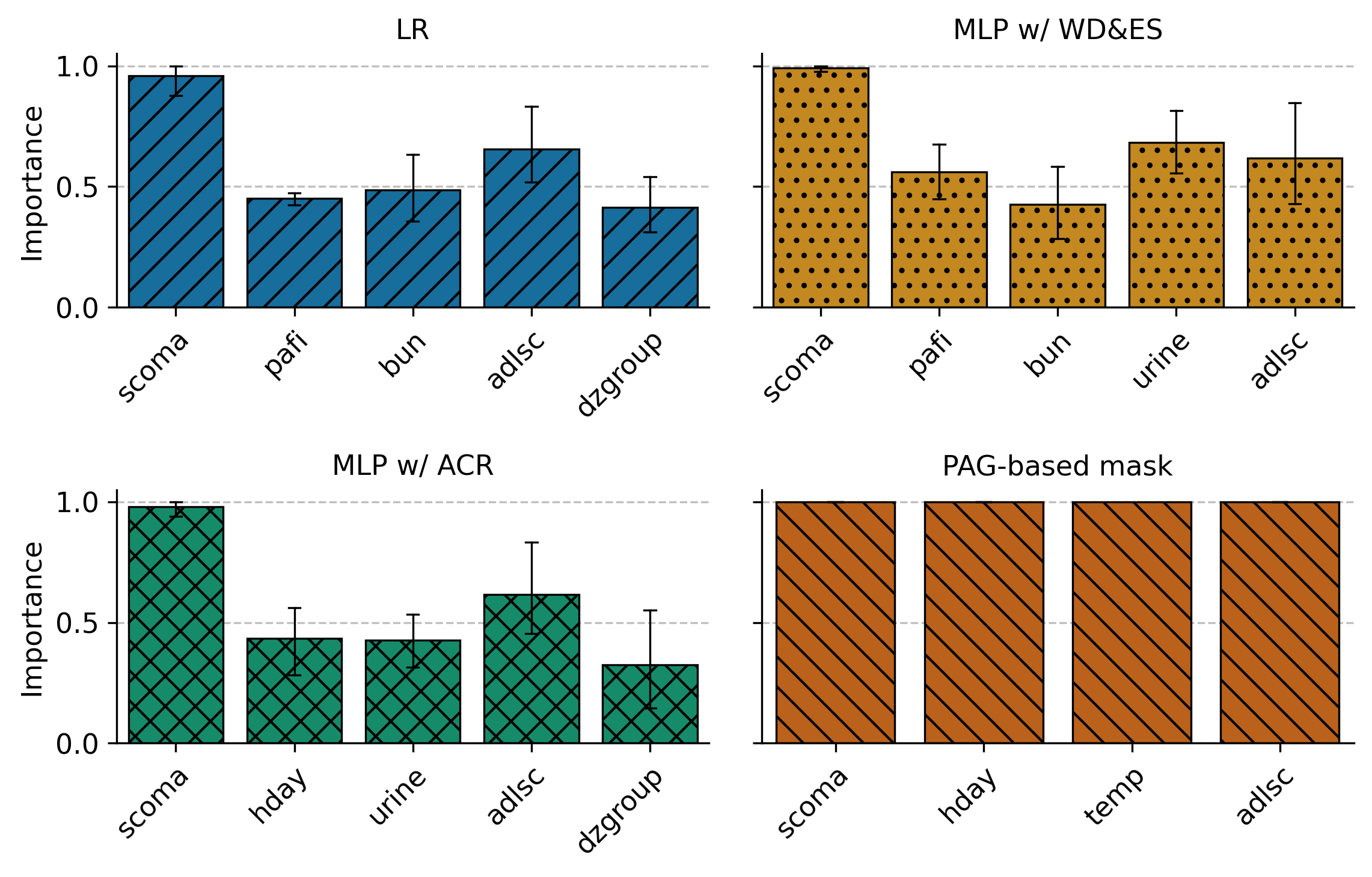}
    \caption{Normalized feature importance scores for top-3 features of each model for the SUPPORT dataset.}
    \label{fig:support_bar}
\end{figure}

To optimize for both predictive accuracy and causal robustness, we set the regularization strength to $\lambda=1e^-2$. We performed five-fold cross-validation and report the test metrics across the folds (Fig.~\ref{fig:support_box}). We also plot the normalized importance scores for the top-5 features of each model (Fig.~\ref{fig:support_bar}). As seen in the figures, MLP w/ ACR achieves competitive predictive accuracy while prioritizing causally robust features. In contrast to MLP w/ WD and ES, MLP w/ ACR picks features like hday (day in hospital at study entry) and dzgroup (patient's disease category). This implies that the model looks into robust proxies for latent patient severity and chronicity to contextualize acute physiological markers (lab values), thereby improving in-hospital mortality predictions.

\section{Discussion}
Accurate prediction of outcomes enhances clinical decision-making and personalized patient care, ultimately improving patient outcomes. Supervised machine learning algorithms are commonly used for outcome prediction in the medical domain. Most modern machine learning algorithms optimize for predictive accuracy, which can result in models latching onto spurious correlations instead of robust predictors. Causal structure learning methods have the potential to provide robust predictors for the target, but can be too conservative because of algorithmic and data assumptions, resulting in loss of diagnostic precision. Therefore, we propose a novel model-agnostic regularization strategy, Adaptive-CaRe, for generalized outcome prediction in the medical domain. Adaptive-CaRe strikes a balance between both predictive value and causal robustness by incorporating a penalty that is proportional to the difference between the estimated statistical contribution and estimated causal contribution of the input features for model predictions. 

Our experiments on synthetic data establishes the efficacy of the proposed Adaptive-CaRe regularizer in finding robust predictors for the target while maintaining competitive predictive accuracy. We demonstrate how models optimized for accuracy, like LR and MLP, favour spurious correlations for predictive accuracy. We also observe that the proposed regularizer consistently outperforms state-of-the-art causal regularizer, CASTLE, in all the experiments because of the conservative nature of CASTLE architecture. Our experiments on varying sample sizes demonstrate the potential of the regularizer in low-data regimes which is common in the medical domain. With experiments on a standard causal benchmark, we provide a blueprint for navigating the trade-off between predictive accuracy and causal robustness by tweaking the regularization strength, $\lambda$. Validation using a real-world dataset confirms that the results translate to practical, real-domain settings.

Therefore, Adaptive-CaRe provides a simple yet effective solution to the long-standing trade-off between predictive accuracy and causal robustness of machine learning models in the medical domain. Future work would involve studying the behaviour of Adaptive-CaRe regularizer with alternate causal structure learning frameworks and complex classification models to provide deeper insights into the outcome prediction problem in the medical domain at a larger scale. 

\appendices
\section{PAG-based causal mask}
~\label{fci_details}
To generate the PAG-based causal mask, we make use of the FCI implementation provided in causal-learn package \cite{zheng2024causal}. As most of the common datasets in the medical domain contain mixed continuous and categorical variables, we utilize Predictive Permutation CIT (PPCIT) \cite{li2024mixed}, a conditional independence test grounded in Predictive Permutation Feature Importance (PPFI) \cite{watson2021testing}. We use the XGBoost classification and regression models \cite{chen2016xgboost} as the non-parametric models inside PPCIT. We set the significance level for the conditional independence tests at $\alpha=0.1$. 

To obtain the PAG-based causal mask, we compute a binary adjacency mask $A \in \{0, 1\}^d$ derived from the PAG. Let $\mu(X_j, Y)$ denote the edge mark at the $X_j$ end of the edge between variable $X_j$ and the target $Y$, and $\mu(Y, X_j)$ denote the mark at the $Y$ end. We assign $A_j = 1$ if and only if $X_j$ is a possible direct cause of $Y$ according to the criteria defined above:

\begin{equation}
    \resizebox{0.48\textwidth}{!}{
    $A_j = 
    \begin{cases} 
    1 & \text{if } \mu(Y, X_j) = \text{Arrow} \quad \text{AND} \quad \mu(X_j, Y) \in \{ \text{Tail}, \text{Circle},  \text{Arrow} \} \\
    0 & \text{otherwise}
    \end{cases}
    $}
    \label{eq:mask}
\end{equation}

This masking procedure effectively filters out edges representing undefined relationships, retaining only those features with either high causal plausibility or high predictive accuracy. 

\section{Normalized feature importance scores}
~\label{shap}
We computed and plotted normalized feature importance scores to interpret the predictions of LR and MLP models. For the MLP models, we utilized SHAP (SHapley Additive exPlanations) values \cite{lundberg2017unified} estimated via the Kernel SHAP method. To reduce computational complexity, the background dataset was summarized using $k$-means clustering with $k=10$ centroids derived from the training distribution. Global feature importance was calculated as the mean absolute SHAP value across a test subset of 50 samples. For the binary classification task, we averaged the absolute contributions across both output classes. Finally, we normalized the resulting importance scores by dividing them by the maximum score, scaling the most significant feature to 1.0. For the LR model, we utilized the Linear SHAP estimator.

\section*{Acknowledgment}
This work is supported by the project "Next Generation AI Computing (gAIn)," funded by the Bavarian Ministry of Science and the Arts and the Saxon Ministry for Science, Culture, and Tourism.

\section*{References}
\bibliographystyle{IEEEtran}
\bibliography{references}

\end{document}